\documentclass{article}

\usepackage[preprint,nonatbib]{nips_2018}

\usepackage[citestyle=numeric,natbib=true]{biblatex}
\usepackage[compact]{titlesec}
\usepackage[utf8]{inputenc} 
\usepackage[T1]{fontenc}    
\usepackage{amsfonts}       
\usepackage{microtype}      
\usepackage{epsfig}
\usepackage{graphicx}
\usepackage{amsmath}
\usepackage{amssymb}
\usepackage{caption}
\usepackage{subcaption}
\usepackage{gensymb}
\usepackage{appendix}

\usepackage{booktabs}
\usepackage{float}
\usepackage{hhline}
\usepackage{tabularx}
\newcolumntype{Y}{>{\centering\arraybackslash}X}
\newcolumntype{s}{>{\hsize=.8\hsize}Y}
\newcolumntype{t}{>{\hsize=.6\hsize}Y}
\newcolumntype{b}{X}
\newcolumntype{?}{!{\vrule width 1pt}}
\usepackage{multirow}
\restylefloat{table}
\usepackage{makecell}
\usepackage{stfloats}
\usepackage{wrapfig}

\makeatletter
\newcommand{\thickhline}{%
    \noalign {\ifnum 0=`}\fi \hrule height 1pt
    \futurelet \reserved@a \@xhline
}
\newcolumntype{"}{@{\hskip\tabcolsep\vrule width 1pt\hskip\tabcolsep}}
\makeatother

\newcommand\NameCite[1]{%
  \citeauthor{#1}~[\citeyear{#1}]}

\usepackage[pdfusetitle,breaklinks=true,colorlinks,citecolor=black,bookmarks=false]{hyperref}
\usepackage{pdfpages}
\usepackage{cleveref}
\crefname{appsec}{Appendix}{Appendices}

\bibliography{biblio}

\title{Benchmarking Neural Network Robustness to\\Common Corruptions and Surface Variations}

\author{Dan Hendrycks\\
University of California, Berkeley\\
{\tt hendrycks@berkeley.edu}
\And
Thomas G. Dietterich\\
Oregon State University\\
{\tt tgd@oregonstate.edu}
}

\begin{document}
\setlength{\abovedisplayskip}{3pt}
\setlength{\belowdisplayskip}{3pt}

\maketitle

\begin{abstract}
In this paper we establish rigorous benchmarks for image classifier robustness. Our first benchmark, \textsc{ImageNet-C}, standardizes and expands the corruption robustness topic, while showing which classifiers are preferable in safety-critical applications. Unlike recent robustness research, this benchmark evaluates performance on commonplace corruptions not worst-case adversarial corruptions. We find that there are negligible changes in relative corruption robustness from AlexNet to ResNet classifiers, and we discover ways to enhance corruption robustness. Then we propose a new dataset called \textsc{Icons-50} which opens research on a new kind of robustness, surface variation robustness. With this dataset we evaluate the frailty of classifiers on new styles of known objects and unexpected instances of known classes. We also demonstrate two methods that improve surface variation robustness. Together our benchmarks may aid future work toward networks that learn fundamental class structure and also robustly generalize.\\
\textbf{This document is superseded by \href{https://arxiv.org/abs/1903.12261}{\texttt{https://arxiv.org/abs/1903.12261}}}
\end{abstract}

\section{Introduction}
\noindent The human vision system is robust in ways that existing computer vision systems are not \citep{transforms,recht}. Unlike current deep learning classifiers~\citep{AlexNet,resnet,resnext}, the human vision system is not fooled by small changes in query images. Humans are also not confused by many forms of corruption such as snow, blur, pixelation, and novel combinations of these. Humans can even deal with abstract changes in structure and style. Achieving these kinds of robustness is an important goal for computer vision and machine learning. It is also essential for creating deep learning systems that can be deployed in safety-critical applications.

Most work on robustness in deep learning methods for vision has focused on the important challenges of robustness to adversarial examples \citep{adversarial,bypass,breakdistill}, unknown unknowns \citep{hendrycks_baseline,pacanomaly,hendrycks2018oe}, and model or data poisoning \citep{byzantine,poison,glc}. In contrast, we develop and validate datasets for two other forms of robustness. Specifically, we introduce the \textsc{ImagetNet-C} dataset for input \emph{corruption robustness}~\citep{igor} and the \textsc{Icons-50} dataset for \emph{surface variation robustness}. These challenges can be overcome by future networks which do not rely on spurious correlations or cues inessential to the object's class.

To create \textsc{ImageNet-C}, we introduce a set of 75 common visual corruptions and apply them to the ImageNet object recognition challenge~\citep{imagenet}. We hope that this will serve as a general dataset for benchmarking robustness to image corruptions and prevent methodological problems such as moving goal posts and result cherry picking. We evaluate the performance of current deep learning systems and show that there is wide room for improvement on \textsc{ImageNet-C}. We also introduce methods and architectures that improve robustness on \textsc{ImageNet-C} without losing accuracy.

We then benchmark surface variation robustness. In this setting the fundamental class structure is unchanged, but the object's surface-level statistics and superficial aspects vary but are not corrupted. To benchmark surface variation robustness, we introduce the \textsc{Icons-50} dataset. This dataset is intended to support research on robustness to surface variations such as the introduction of new styles and novel animal species. We then describe two methods that improve surface variation robustness. By defining and benchmarking surface variation robustness and corruption robustness, we facilitate research that leads to classifiers that track the fundamental structure of the classes and robustly generalize to unexpected inputs.

\section{Related Work}

\noindent\textbf{Adversarial Examples.}\quad An adversarial image is a clean image perturbed by a small corruption so as to confuse a classifier. These deceptive corruptions can occasionally fool black-box classifiers \citep{kurakin}. Algorithms have been developed that search for the smallest corruption in RGB space that is sufficient to confuse a classifier~\citep{ground}. Thus adversarial corruptions serve as type of worst-case analysis for network robustness. Its popularity has often led ``adversarial robustness'' to become interchangeable with ``robustness'' in the literature \citep{bastini, foolbox}.
In efforts to ground adversarial robustness research,
several competing robustness measures have been introduced \citep{bastini,carlinimetric,ground,badmitpaper,katz}. New defenses \citep{forsyth, distill, defense, defense2} quickly succumb to new attacks~\citep{contraforsyth,bypass,breakdistill}.

\noindent\textbf{Robustness in Speech.}\quad Speech recognition research emphasizes robustness to common corruptions over worst-case, adversarial corruptions~\citep{overviewmicrosoft,cmuoverview}. Common acoustic corruptions (e.g., street noise, background chatter, wind) receive greater focus than adversarial audio, because common corruptions are ever-present and unsolved. There are several popular datasets containing noisy test audio \citep{aurora2,aurora5}. Robustness in noisy environments requires robust architectures, and some research finds convolutional networks more robust than fully connected networks~\citep{convvsfcn}. Additional robustness has been achieved through pre-processing techniques such as standardizing the statistics of the input \citep{cmn,histo,histospeech,pncc}.

\noindent\textbf{ConvNet Fragility Studies.}\quad Several studies demonstrate the fragility of convolutional networks on simple corruptions. For example, \NameCite{impulse} use impulse noise to break Google's Cloud Vision API. Using Gaussian noise and blur,~\NameCite{dodgeeval} demonstrate the superior robustness of human vision to convolutional networks, \emph{even after networks are fine-tuned} on Gaussian noise or blur.~\NameCite{eidelon} compare networks to humans on noisy and elastically deformed images. They find that fine-tuning on specific corruptions does not generalize, and classification error patterns underlying network and human predictions are not similar. Others show that networks fail to generalize to images deformed by slight geometric transformations~\citep{transforms}.

\noindent\textbf{Robustness Enhancements.}\quad In an effort to reduce classifier fragility, \NameCite{igor} fine-tune on blurred images. They find it is not enough to fine-tune on one type of blur to generalize to other blurs. Furthermore, fine-tuning on several blurs can marginally decrease performance. \NameCite{stability} also find that fine-tuning on noisy images can cause underfitting, so they encourage the noisy image softmax distribution to match the clean image softmax. \NameCite{dodge2} address underfitting via an ensemble. They fine-tune each network on one corruption and classify with an mixture of these corruption-specific experts, though they do not assess combinations of known corruptions.

\section{The \textsc{ImageNet-C} Corruption Robustness Benchmark}

\subsection{The \textsc{ImageNet-C} Dataset}
\begin{figure}[tp]
  \begin{center}
      \includegraphics[width=\textwidth]{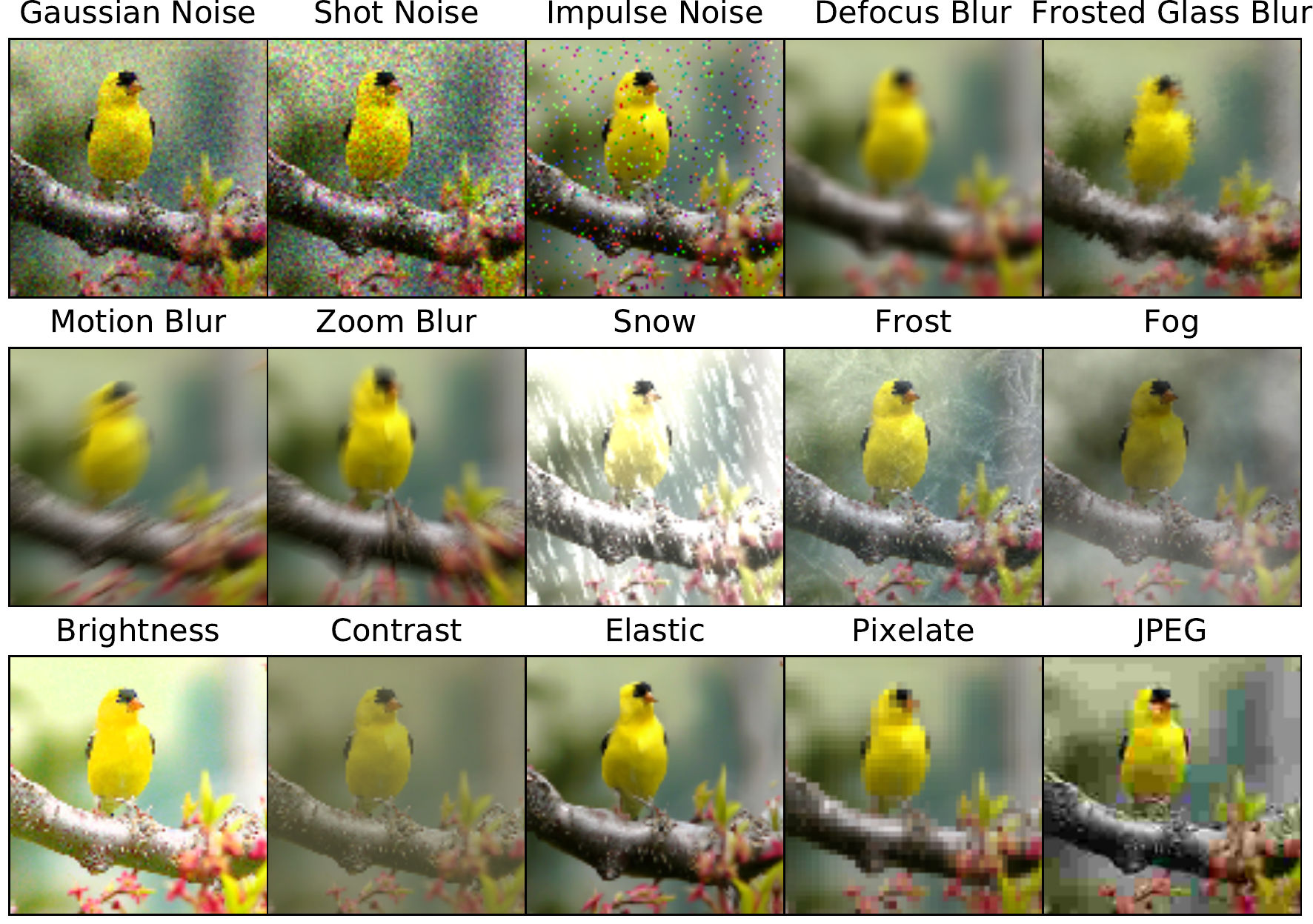}
  \end{center}
  \caption{Our \textsc{ImageNet-C} dataset consists of 15 types of algorithmically generated corruptions from noise, blur, weather, and digital categories. Each type of corruption has five levels of severity, resulting in 75 distinct corruptions. See different severity levels in the Supplementary Materials.}
\label{fig:corruptedbird}
\vspace{-5pt}
\end{figure}

\noindent\textbf{\textsc{ImageNet-C} Design.}\quad Our corruption robustness benchmark consists of 15 diverse corruption types, exemplified in Figure~\ref{fig:corruptedbird}. The benchmark covers noise, blur, weather, and digital categories. Research that improves performance on this benchmark should indicate general robustness gains, as the corruptions are varied and great in number. These 15 corruption types each have five different levels of severity, since corruptions can manifest themselves at varying intensities. The Supplementary Materials gives an example of a corruption type's five different severities. Real-world corruptions also have variation even at a fixed intensity. To simulate these, we introduce variation for each corruption when possible. For example, each fog cloud is unique to each image. These algorithmically generated corruptions are applied to the ImageNet~\citep{imagenet} validation images to produce our corruption robustness dataset \textsc{ImageNet-C}. The dataset can be downloaded or re-created by visiting \href{https://github.com/hendrycks/robustness}{\texttt{https://github.com/hendrycks/robustness}}. Our benchmark tests networks with \textsc{ImageNet-C} images, \emph{but networks do not train on these images}. Networks are trained with datasets such as ImageNet but not \textsc{ImageNet-C}. To enable further experimentation, we also designed an extra corruption for each noise, blur, weather, or digital category. Extra corruptions are depicted and explicated in the Supplementary Materials and also available at the aforementioned URL. Even more experimentation is possible by applying these distortions to other images, such as those from Places365. We find that applying these corruptions to $32\times32$ images and benchmarking on such small images is not predictive of corruption robustness on larger images. Overall, the dataset \textsc{ImageNet-C} consists of 15 corruption types, each with five different severities, all applied to ImageNet validation images for testing a pre-existing network.

\noindent\textbf{Common Corruptions.}\quad The first \textsc{ImageNet-C} corruption is \emph{Gaussian noise}. This corruption can appear in low-lighting conditions. \emph{Shot noise}, also called Poisson noise, is electronic noise caused by the discrete nature of light itself. \emph{Impulse noise} is a color analogue of salt-and-pepper noise and can be caused by bit errors. \emph{Defocus blur} occurs when an image is out of focus. \emph{Frosted Glass Blur} appears with ``frosted glass'' windows or panels. \emph{Motion blur} appears when a camera is moving quickly. \emph{Zoom blur} occurs when a camera moves toward an object rapidly. \emph{Snow} is a visually obstructive form of precipitation. \emph{Frost} forms when lenses or windows are coated with ice crystals. \emph{Fog} shrouds objects and is rendered with the diamond-square algorithm. \emph{Brightness} varies with daylight intensity. \emph{Contrast} can be high or low depending on lighting conditions and the photographed object's color. \emph{Elastic} transformations stretch or contract small image regions. \emph{Pixelation} occurs when upsampling a low-resolution image. \emph{JPEG} is a lossy image compression format that increases image pixelation and introduces artifacts. Each corruption type is tested with depth due to its five severity levels, and this broad range of corruptions allows us to test model corruption robustness with breadth.

\subsection{Metric and Setup}
\noindent\textbf{Mean and Relative Corruption Error.}\quad
Common corruptions such as Gaussian noise can be benign or destructive depending on their severity. In order to \emph{comprehensively} evaluate a classifier's robustness to a given type of corruption, we score the classifier's performance across five corruption severity levels and aggregate its scores. The first evaluation step is to take a pre-existing classifier here notated ``Network,'' which has not and will not train on \textsc{ImageNet-C}, and then compute the clean dataset top-1 error rate. Denote this error rate $E^\text{Network}_\text{Clean}$. This same classifier will then test on an \textsc{ImageNet-C} corruption type notated ``Corruption.'' Let top-1 error rate for the Network classifier on Corruption with severity level $s$ ($1\le s \le 5$) be written $E_{s,\text{Corruption}}^\text{Network}$. The classifier's aggregate performance across the five severities of the corruption type Corruption is the Corruption Error, computed with the formula
\[ \text{CE}_\text{Corruption}^\text{Network} = \sum_{s=1}^5 E_{s,\text{Corruption}}^\text{Network}\bigg/\sum_{s=1}^5 E_{s,\text{Corruption}}^\text{AlexNet}. \]
Different corruptions pose different levels of difficulty. For example, fog corruptions often obscure an object's class more than Brightness corruptions. Thus to make Corruption Errors comparable across corruption types, we adjust for the difficulty by dividing by AlexNet's errors. Now with commensurate Corruption Errors, we can summarize model corruption robustness by averaging the 15 Corruption Error values $\left(\text{CE}_\text{Gaussian Noise}^\text{Network}, \text{CE}_\text{Shot Noise}^\text{Network}, \ldots, \text{CE}_\text{JPEG}^\text{Network}\right)$. This results in the \emph{mean CE} or \emph{mCE} for short.

We now introduce a more nuanced corruption robustness measure. Consider a classifier that withstands most corruptions, so that the gap between the mCE and the clean data error is minuscule. Contrast this with a classifier with a low clean error rate that does not cope well with corruptions, which corresponds to a large gap between the mCE and clean data error. It is possible the former classifier has a larger mCE than the latter, despite the former degrading more gracefully in the presence of corruptions. The amount that the classifier declines on corrupted inputs is given by the formula
\[
\text{Relative CE}_\text{Corruption}^\text{Network} = \bigg(\sum_{s=1}^5 E_{s,\text{Corruption}}^\text{Network} - E^\text{Network}_\text{Clean}\bigg)\bigg/\bigg(\sum_{s=1}^5 E_{s,\text{Corruption}}^\text{AlexNet} - E^\text{AlexNet}_\text{Clean}\bigg).
\]
Averaging these 15 Relative Corruption Errors results in the \emph{Relative mCE}. In short, the Relative mCE measures the relative robustness or the performance degradation when encountering corruptions.

\begin{wrapfigure}{r}{0.5\textwidth}
\vspace{-10pt}
  \begin{center}
      \includegraphics[width=0.48\textwidth]{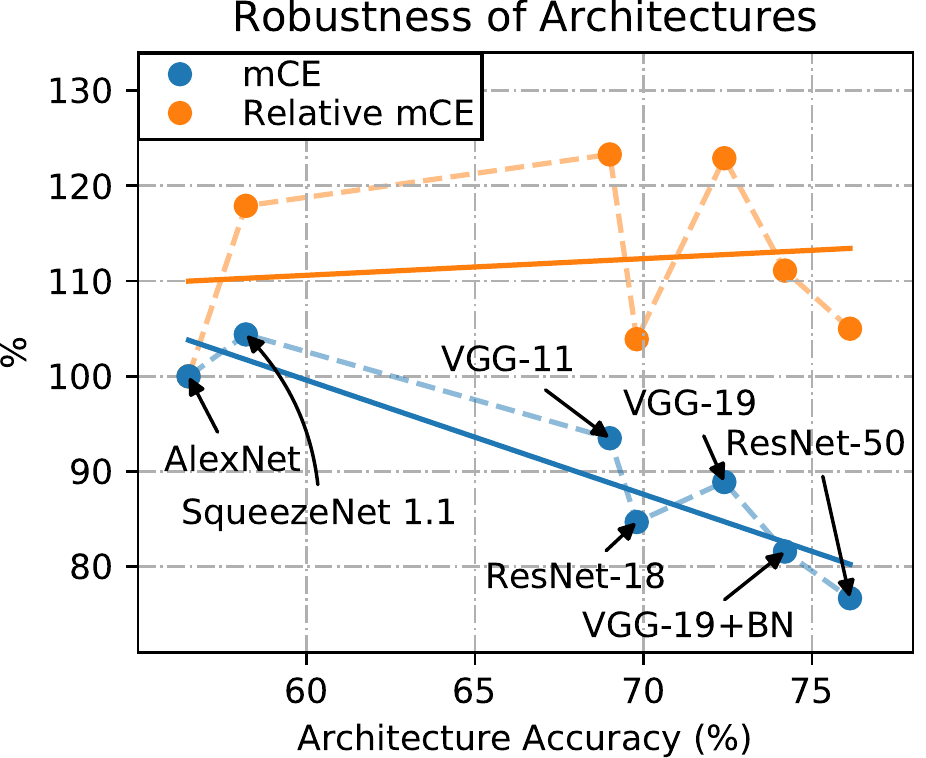}
  \end{center}
  \caption{Robustness (mCE) and Relative mCE \textsc{ImageNet-C} values. Relative mCE values suggest robustness in itself declined from AlexNet to ResNet. ``BN'' abbreviates Batch Normalization.}
\label{fig:architectureerrors}
\vspace{-15pt}
\end{wrapfigure}

\noindent\textbf{Preserving Metric Validity.}\quad
The goal of ImageNet-C is to evaluate the robustness of machine learning algorithms to novel forms of corruption. Humans are able to generalize to novel corruptions quite well. For example, they can easily deal with new Instagram filters. Hence, we propose the following protocol. The image recognition network should be trained on the ImageNet training set and on whatever other training sets the investigator wishes to include. However, the network should not be trained on any of the 75 corruptions that were used to generate \textsc{ImageNet-C}. Then the resulting trained model should be evaluated on \textsc{ImageNet-C} and the above metrics computed. We provide a separate set of additional corruptions that can be employed for training or validation prior to testing on \textsc{ImageNet-C} (see Supplementary Materials). 


\subsection{Architecture Robustness}

\noindent How robust are current methods and has progress in computer vision been achieved at the expense of robustness? As seen in Figure~\ref{fig:architectureerrors}, as architectures improve, so too does the mean Corruption Error (mCE). By this measure, architectures have become progressively more successful at generalizing to corrupted distributions. All Corruption Error values are in~\ref{tab:archi}. Note that models with similar clean error rates have fairly similar CEs, and there are no large shifts in a corruption type's CE. Consequently, it would seem that architectures have slowly and consistently improved their representations over time.
However, it appears that robustness improvements are mostly explained by accuracy improvements. Recall that the Relative mCE tracks a classifier's accuracy decline in the presence of corruptions. Figure~\ref{fig:architectureerrors} shows that the Relative mCE is worse than that of AlexNet~\citep{AlexNet}. In consequence, from AlexNet to ResNet~\citep{resnet}, robustness in itself has barely changed. Relative robustness remains near AlexNet-levels and therefore below human-level, which shows that our ``superhuman'' classifiers are decidedly subhuman. We now apply \textsc{ImageNet-C} to evaluate several methods for attempting to improve robustness to image corruption.

\begin{table}[ht]
\footnotesize
\begin{center}
{\setlength\tabcolsep{1.3pt}%
\begin{tabular}{@{}l c ? c | c c c | c c c c | c c c  c | c c c c@{}}
\multicolumn{3}{c}{} & \multicolumn{3}{c}{Noise} & \multicolumn{4}{c}{Blur} & \multicolumn{4}{c}{Weather} & \multicolumn{4}{c}{Digital} \\
\cline{1-18}
Network & Error & \multicolumn{1}{c|}{\,\textbf{mCE}\,} & \scriptsize{Gauss.}
    & \scriptsize{Shot} & \scriptsize{Impulse} & \scriptsize{Defocus} & \scriptsize{Glass} & \scriptsize{Motion} & \scriptsize{Zoom} & \scriptsize{Snow} & \scriptsize{Frost} & \scriptsize{Fog} & \scriptsize{Bright} & \scriptsize{Contrast} & \scriptsize{Elastic} & \scriptsize{Pixel} & \scriptsize{JPEG}\\ \hline 
AlexNet    & 43.5 & 100.0 & 100 & 100 & 100 & 100 & 100 & 100 & 100 & 100 & 100 & 100 & 100 & 100 & 100 & 100 & 100 \\
SqueezeNet & 41.8 & 104.4 & 107 & 106 &	105 & 100 &	103 & 101 &	100 & 101 & 103 & 97  & 97  & 98  & 106 & 109 & 134 \\
VGG-11     & 31.0 & 93.5  & 97  &  97 & 100 & 92  & 99  & 93  & 91  & 92  & 91  & 84  & 75  & 86  & 97  & 107 & 100 \\
VGG-19     & 27.6 & 88.9  & 89  & 91  & 95  & 89  & 98  & 90  & 90  & 89  & 86  & 75  & 68  & 80  & 97  & 102 & 94 \\
VGG-19+BN  & 25.8 & 81.6  & 82  & 83  & 88  & 82  & 94  & 84  & 86  & 80  & 78  & 69  & 61  & 74  & 94  & 85  & 83 \\
ResNet-18  & 30.2 & 84.7  & 87  & 88  & 91  & 84  & 91  & 87  & 89  & 86  & 84  & 78  & 69  & 78  & 90  & 80  & 85 \\
ResNet-50  & 23.9 & 76.7  & 80  & 82  & 83  & 75  & 89  & 78  & 80  & 78  & 75  & 66  & 57  & 71  & 85  & 77  & 77 \\
\Xhline{2\arrayrulewidth}
\end{tabular}}
\caption{Corruption Error and mCE values of different corruptions and architectures on \textsc{ImageNet-C}. The mCE value is the mean Corruption Error of the corruptions in Noise, Blur, Weather, and Digital columns. All models are trained on clean ImageNet images, not \textsc{ImageNet-C} images. Here ``BN'' abbreviates Batch Normalization~\citep{bn}.
}
\label{tab:archi}
\end{center}
\vspace{-10pt}
\end{table}

\subsection{Informative Robustness Enhancement Attempts}
\noindent \textbf{Stability Training.}\quad Stability training is a technique to improve the robustness of deep networks~\citep{stability}. The method's creators found that training on images corrupted with noise can lead to underfitting, so they instead propose minimizing the cross-entropy from the noisy image's softmax distribution to the softmax of the clean image. The authors evaluated performance on images with subtle differences and suggested that the method provides additional robustness to JPEG corruptions. We fine-tune a ResNet-50 with stability training for five epochs. For train time corruptions, we corrupt images with uniform noise, where the maximum and minimum of the uniform noise is tuned over $\{0.01, 0.05, 0.1\}$, and the stability weight is tuned over $\{0.01, 0.05, 0.1\}$. Across all noise strengths and stability weight combinations, the models with stability training tested on \textsc{ImageNet-C} had a larger mCEs than the baseline ResNet-50's mCE. Even on unseen noise corruptions, stability training did not increase robustness. An upshot of this failure is that benchmarking robustness-enhancing techniques requires a diverse test set.

\noindent \textbf{Image Denoising.}\quad An approach orthogonal to modifying model representations is to improve the inputs using image restoration techniques. Although \emph{general} image restoration techniques are not yet mature, denoising restoration techniques are not. We thus attempt restore an image with the denoising technique called non-local means~\citep{nlmeans}. The amount of denoising applied is determined by the noise estimation technique of~\NameCite{idealspatial}. Therefore clean images receive nearly no modifications from the restoration method, while noisy images should undergo considerable restoration. We found that denoising increased the mCE from $76.7\%$ to $82.1\%$. A plausible account is that the non-local means algorithm slightly smoothed images even when images lacked noise, despite having the non-local means algorithm governed by the noise estimate. Therefore, the gains in noise robustness were wiped away by subtle blurs to images with other types of corruptions, showing that targeted image restoration can prove harmful for robustness.

\noindent \textbf{Smaller Models.} All else equal, ``simpler'' models often generalize better, and ``simplicity'' frequently translates to model size. Accordingly, smaller models may be more robust. We test this hypothesis with CondenseNets~\citep{condensenet}. A CondenseNet attains its small size via sparse convolutions and pruned filter weights. An off-the-shelf CondenseNet ($C=G=4$) obtains a 26.3\% error rate and a 80.8\% mCE. On the whole, this CondenseNet is slightly less robust than larger models of similar accuracy. Even more pruning and sparsification yields a CondenseNet ($C=G=8$) with both deteriorated performance (28.9\% error rate) and robustness (84.6\% mCE). Here again robustness is worse than larger model robustness. Though models fashioned for mobile devices are smaller and in some sense simpler, this does not improve robustness.

\subsection{Successful Corruption Robustness Enhancements}
\noindent \textbf{Histogram Equalization.}\quad Histogram equalization successfully standardizes speech data for robust speech recognition~\citep{histo,histospeech}. For images, we find that preprocessing with Contrast Limited Adaptive Histogram Equalization~\citep{clahe} is quite effective. Unlike our previous image denoising attempt, CLAHE reduces the effect of some corruptions while not worsening performance on most others, thereby improving the mCE. We demonstrate CLAHE's net improvement by taking a pre-trained ResNet-50  and fine-tuning the whole model for five epochs on images processed with CLAHE. The ResNet-50 has a 23.87\% error rate, but ResNet-50 with CLAHE has an error rate of 23.55\%. On nearly all corruptions, CLAHE slightly decreases the Corruption Error. The ResNet-50 without CLAHE preprocessing has an mCE of 76.7\%, while with CLAHE the ResNet-50's mCE decreases to 74.5\%.

\begin{figure}
    \centering
    \begin{minipage}{.48\textwidth}
      \centering
      \includegraphics[width=\linewidth]{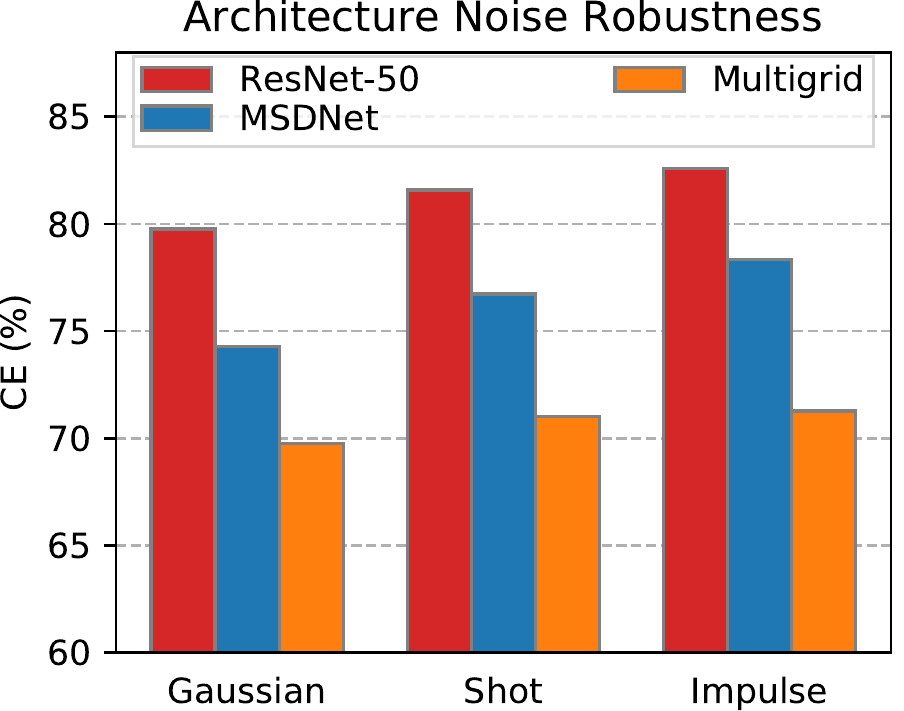}
      \caption{Architectures like Multigrid networks and DenseNets resist noise corruptions more effectively than ResNets.}
    \label{fig:multigrid}
    \end{minipage}\hfill%
        \begin{minipage}{.48\textwidth}
      \centering
      \includegraphics[width=\linewidth]{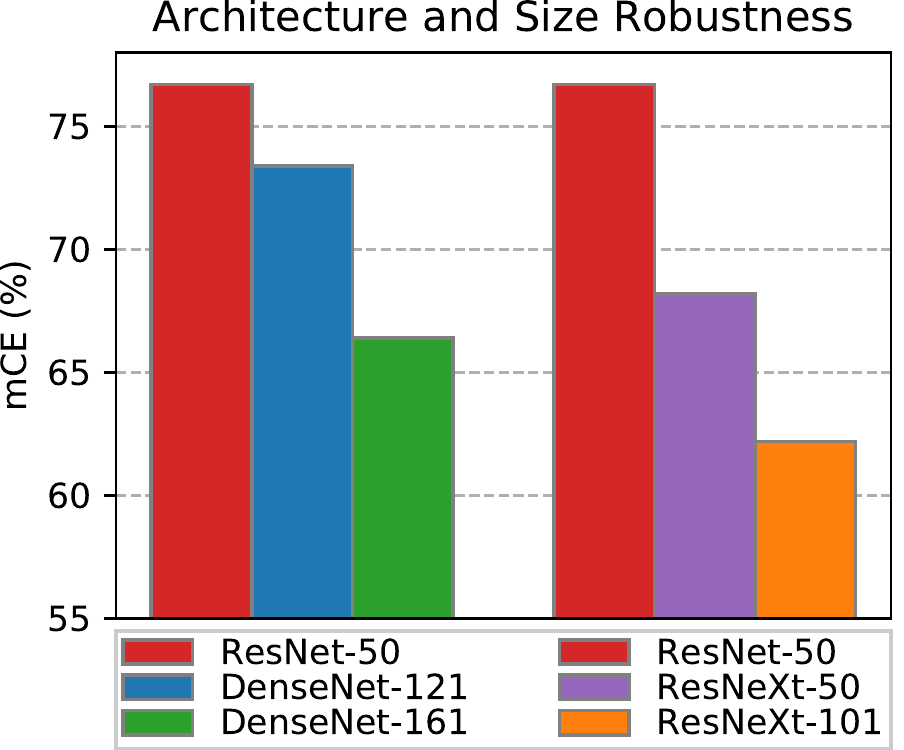}
      \caption{Larger feature aggregating networks achieve robustness gains that substantially outpace their accuracy gains.}
    \label{fig:bigmodels}
    \end{minipage}\hfill
\vspace{-5pt}
\end{figure}

\noindent \textbf{Multiscale, Feature Aggregating, and Larger Networks.}\quad
Multiscale architectures achieve greater robustness by propagating features across scales at each layer rather than slowly gaining a global representation of the input as in typical convolutional neural networks. Some multiscale architectures are called Multigrid Networks~\citep{multigrid}. Multigrid networks each have a pyramid of grids in each layer which enables the subsequent layer to operate across scales. Along similar lines, Multi-Scale Dense Networks (MSDNets)~\citep{MSDNet} use information across scales. MSDNets bind network layers with DenseNet-like~\citep{densenet} skip connections. These two different multiscale networks both enhance robustness. Before comparing mCE values, we first note the Multigrid network has 24.6\% top-1 error, as does the MSDNet, while the ResNet-50 has 23.9\% top-1 error. On noisy inputs, Multigrid networks noticably surpass ResNets and MSDNets, as shown in Figure~\ref{fig:multigrid}. Since multiscale architectures have high-level representations processed in tandem with fine details, the architectures appear better equipped to suppress otherwise distracting pixel noise. When all corruptions are evaluated, ResNet-50 has an mCE of 76.7\%, the MSDNet has an mCE of 73.6\%, and the Multigrid network has an mCE of 73.3\%.

Some recent models enhance the ResNet architecture by increasing what is called feature aggregation. Of these, DenseNets~\citep{densenet} and ResNeXts~\citep{resnext} are most prominent. Each purports to have stronger representations than ResNets, and the evidence is largely a hard-won ImageNet error-rate downtick. Interestingly, the \textsc{ImageNet-C} mCE clearly indicates that DenseNets and ResNeXts have superior representations. Accordingly, a switch from a ResNet-50 (23.9\% top-1 error) to a DenseNet-121 (25.6\% error) decreases the mCE from 76.7\% to 73.4\%. More starkly, switching from a ResNet-50 to a ResNeXt-50 (22.9\% top-1) drops the mCE from \emph{76.7\% to 68.2\%}. Results are summarized in Figure~\ref{fig:bigmodels}. This shows that corruption robustness may be a better way to measure future progress in object recognition than the clean dataset top-1 error rate.

Some of the greatest and simplest robustness gains sometimes emerge from making recent models more monolithic. Apparently more layers, more connections, and more capacity allow these massive models to operate more stably on corrupted inputs. We saw earlier that making models smaller does the opposite.  Swapping a DenseNet-121 (25.6\% top-1) with the larger DenseNet-161 (22.9\% top-1) decreases the mCE from 73.4\% to 66.4\%. In a similar fashion, a ResNeXt-50 (22.9\% top-1) is less robust than the a giant ResNeXt-101 (21.0\% top-1); the mCEs are 68.2\% and 62.2\% respectively. Both model size and feature aggregation results are summarized in Figure~\ref{fig:bigmodels}. Consequently, future models with more depth, width, and feature aggregation may attain further robustness.

\section{Surface Variation Robustness}

\begin{figure}[tp]
  \begin{center}
      \includegraphics[width=\textwidth]{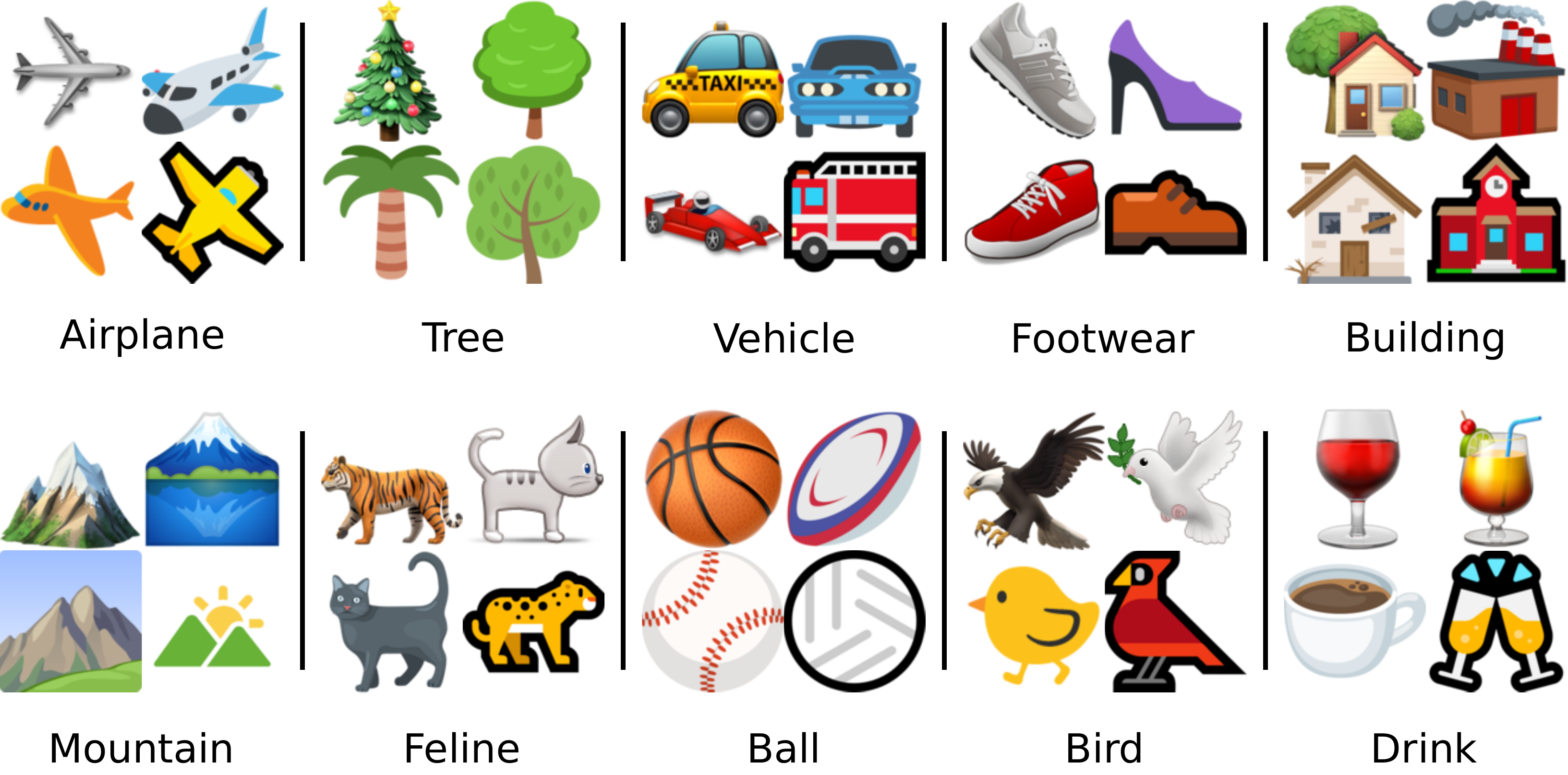}
  \end{center}
  \caption{Image samples are from 12 of the 50 \textsc{Icons-50} classes. These images showcase the dataset's high image quality, interclass and intraclass diversity, and its many styles.}
\label{fig:icons}
\end{figure}

\noindent An important goal for machine learning is to learn the essential structural elements of a class while being robust to unimportant surface variation. Whereas corruptions degrade the entirety of an image, surface variations cleanly alter the surface of an object and preserve the fundamental structure of the class. We developed the \textsc{Icons-50} dataset to test robustness to two forms of surface variation: style variation and subtype variation. Style variation is variation in artistic style that leaves the fundamental structure unchanged. Subtype variation is variation within a broad class (e.g., types of cats within the broad class of ``cat'') where the subtypes all share essential characteristics of the broad class.


\subsection{The \textsc{Icons-50} Dataset}
\noindent The \textsc{Icons-50} dataset consists of 10,000 images belonging to 50 classes of icons (e.g., people, food, activities, places, objects, symbols, etc.) collected from different technology companies and platforms (e.g., Apple, Samsung, Google, Facebook, etc.). A full list of \textsc{Icons-50} classes is in the Supplementary Materials, and the dataset can be downloaded at \href{https://github.com/hendrycks/robustness}{\texttt{https://github.com/hendrycks/robustness}}. A subset of \textsc{Icons-50} classes is shown in Figure~\ref{fig:icons}. Each class has icons with different styles. For instance, icons with the thick, black outlines (like the bottom right ``Drink'' icon) are created by Microsoft. Other styles in the \textsc{Icons-50} dataset are from Apple, Samsung, Google, Facebook, and other platforms. The \textsc{Icons-50} dataset provides more icons per class (mean of 200) and more classes (50) than the logo datasets of \NameCite{flickr} (32 classes, 70 logos per class) and of \NameCite{belga} (37 classes and 53 logos per class).



We propose two protocols for evaluating surface variation robustness using the \textsc{Icons-50} dataset. For style robustness, the icons from one source (e.g., Microsoft) should be held out and the network trained on the remaining icons. Then the ability to generalize to the held out source can be measured by the classification accuracy of the held out icon source. For subtype robustness, we construct a training set by holding out 50 subtypes from \textsc{Icons-50}. For example, we hold out the ``duck'' icons from the broad ``Bird'' class. Entire broad classes are not withheld, only a fraction of the subtypes within \textsc{Icon-50}'s 50 broad classes are withheld. The network is trained on the remaining icons and learns to classify broad classes. We evaluate the classifier's subtype robustness by computing the broad class accuracy on the held out subtypes. These experiments are in the following section, and auxiliary experiments are in the Supplementary Materials.

\subsection{Surface Variation Robustness Experiments with \textsc{Icons-50}}
\noindent \textbf{Style Robustness.}\quad The \textsc{Icons-50} dataset has several styles like Microsoft's flat vector graphics style and Apple's realistic style. We aim to test the model's robustness to an unseen style. To that end, we train a network on \textsc{Icons-50} while holding out Microsoft-styled icons. Microsoft-styled icons appear in all 50 classes, so each class is tested. Here, the metric for style robustness is simply the classifier's accuracy on Microsoft-styled icons after the classifier trains on all other styles. That said, we train a 20 layer ResNet, a 40 layer DenseNet ($L=40,k=24$), and a 29 layer ResNeXt ($8\times32$d) for 50 epochs with the cosine learning rate schedule~\citep{cosine} and Nesterov momentum. Icons are resized to $32\times32$ images. Like others, we use image mirroring and image cropping data augmentation. What we find is that the networks lacks style robustness---the ResNet only obtains 58.3\% accuracy on the held out Microsoft-styled icons, and the DenseNet and ResNeXt do worse with 48.5\% and 51.8\% accuracy, respectively. Clearly DenseNets and ResNeXts do not necessarily outperform ResNets as they did in the corruption robustness benchmark.

These results are not symptomatic of a training data shortage, rather a lack of style robustness. The networks perform well if we instead use \textsc{Icons-50} for traditional classification rather than style robustness. To do this, we hold out one version or rendition of icons that have multiple versions. For example, we hold out one version of the many Google-stylized ``football'' icons which have appeared across different versions of Android. Then we test on these held out versions.
By holding out and testing on one version for icons with many versions, the ResNeXt achieves 98.0\% accuracy, the DenseNet obtains 97.3\% accuracy, and the ResNet 97.2\% accuracy. Thus the networks have enough data. Thus we can conclude that with sub-60\% accuracy on held out Microsoft-styled icons, the networks demonstrate a clear lacking in style robustness.

\noindent \textbf{Subtype Robustness.}\quad Each \textsc{Icons-50} class has many subtypes, so we can treat each class as a broad class. We hold out 50 subtypes, train the classifier on the remaining subtypes to predict the broad class, and test the classifier's accuracy on the held out subtypes. Then, we train a ResNet, DenseNet, and ResNeXt with the same training scheme from the style robustness experiment. After training, we find that classifier accuracies on these held out subtypes are again meager and that style robustness and subtype robustness indeed test different forms of robustness. This time the ResNet is worst, as it obtains only 57.5\% accuracy on the held out subtypes. Meanwhile the DenseNet has 58.7\% accuracy, and the ResNeXt has 60.0\% accuracy. This indicates that the classifiers have wide room for subtype robustness improvement.

\noindent \textbf{Enhancing Surface Variation Robustness.}\quad
Multiscale networks can improve surface variation robustness not just corruption robustness. MSDNets \citep{MSDNet} can process the whole image structure early in the forward pass, so that when the style changes while the fundamental structure is preserved, MSDNets can persevere better than other networks. In comparison to DenseNets, the network type most similar to MSDNets as both use dense connections, the MSDNet obtains 65.1\% accuracy on the style robustness task while the DenseNet obtains only 48.5\% accuracy. However, for subtype robustness, both networks differ by only a fraction of a percent in performance.

A method to improve both aspects of surface variation robustness is Shake-Shake regularization \citep{shake}. Shake-Shake regularization stochastically modulates the influence of each ResNeXt branch, so that the network can gain fault tolerance and endure unusual representations. Note we could not test shake-shake regularization's effect on corruption robustness since contemporary GPUs do not have enough memory for shake-shake regularization applied to networks processing large-scale images. Now, the vanilla ResNeXt has 51.8\% style robustness accuracy, but a ResNeXt with shake-shake regularization jumps to 76.0\% accuracy. Subtype robustness also improves appreciably; the ResNeXt has 60.0\% accuracy while the ResNeXt with shake-shake regularization has 63.6\% accuracy. Clearly subtype robustness can be harder to improve than style robustness, but shake-shake regularization can improve both.

\section{Conclusion}
\noindent In this paper, we introduced what are to our knowledge the first benchmarks for corruption robustness and surface variation robustness. This was made possible by introducing two new datasets, \textsc{ImageNet-C} and \textsc{Icons-50}, the first of which showed that many years of architectural advancements corresponded to minuscule changes in relative robustness. Therefore benchmarking and improving robustness deserves attention, especially as top-1 clean ImageNet accuracy nears its ceiling. We found that some methods harm corruption robustness, while methods such as histogram equalization, multiscale architectures, and larger models improve corruption robustness. Afterward, we opened research in surface variation robustness by defining style and subtype robustness. Here we found modern models to be fragile, but we also found that multiscale networks and highly regularized networks can noticeably enhance surface variation robustness. In this work, we had several findings, introduced novel experiments, and created new datasets for the rigorous study of model robustness, a pressing necessity as models are unleashed into safety-critical real-world settings.

\newpage

{\small
\printbibliography
}

\newpage
\begin{appendices}
\crefalias{section}{appsec}

\section{Example of \textsc{ImageNet-C} Severities}\label{appendixserverity}
\begin{figure}[h]
  \begin{center}
      \includegraphics[width=1\textwidth]{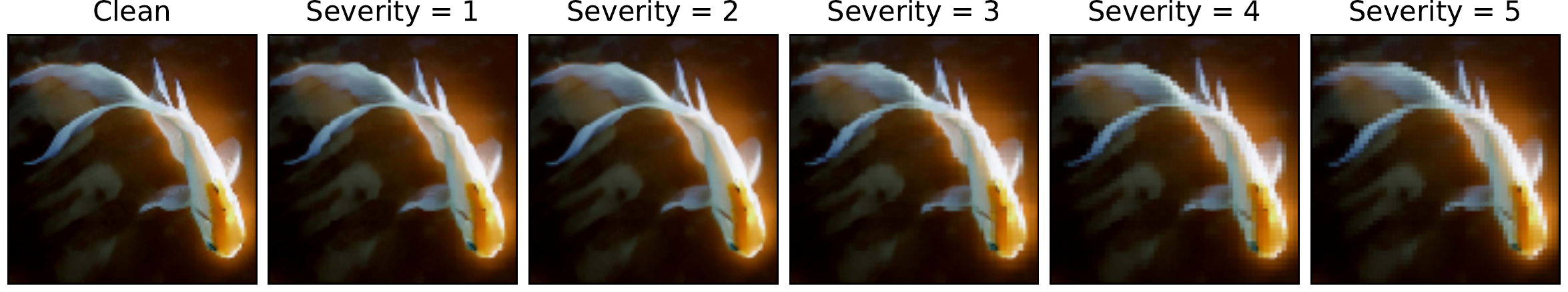}
  \end{center}
  \caption{Pixelation modestly to markedly corrupts a fish, showing our benchmark's varying severities.}
\label{fig:severities}
\end{figure}
\noindent In Figure~\ref{fig:severities}, we show the Pixelation corruption type in its five different severities. Clearly, \textsc{ImageNet-C} corruptions can range from negligible to pulverizing. Because of this range, the benchmark comprehensively assesses each corruption type.

\section{Extra \textsc{ImageNet-C} Corruptions}\label{appendixiconsclassesxtra}
\begin{figure}[h]
  \begin{center}
      \includegraphics[width=1\textwidth]{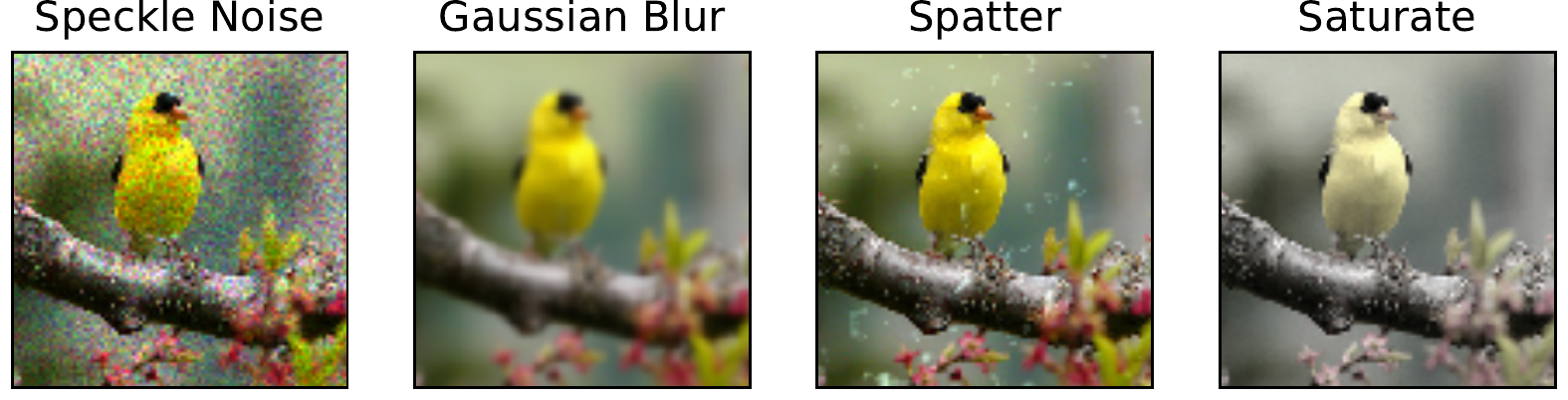}
  \end{center}
  \caption{Extra \textsc{ImageNet-C} corruption examples are available for model validation and sounder experimentation.}
\label{fig:extracorruptions}
\end{figure}
\noindent Directly fitting the types of \textsc{ImageNet-C} corruptions is worth avoiding, as it would cause researchers to overestimate a model's robustness. Therefore, it is incumbent on us to simplify model validation. For this reason, we provide extra corruptions that are available for download at \href{https://github.com/hendrycks/robustness}{\texttt{https://github.com/hendrycks/robustness}}. There is one corruption type for each noise, blur, weather, and digital category. The first corruption type is \emph{speckle noise}, an additive noise where the noise added to a pixel tends to be larger if the original pixel intensity is larger. \emph{Gaussian blur} is a low-pass filter where a blurred pixel is a result of a weighted average of its neighbors, and farther pixels have decreasing weight. \emph{Spatter} can occlude a lens in the form of rain or mud. Finally, \emph{saturate} is common in edited images where images are made more or less colorful. See Figure~\ref{fig:extracorruptions} for instances of each corruption type.

\section{Full Corruption Robustness Results}\label{appendixfull}
\textsc{ImageNet-C} corruption relative robustness results are in Table~\ref{tab:relarchi}. Since we use AlexNet errors to normalize Corruption Error values, we now specify the value $\frac{1}{5}\sum_{s=1}^5 E^\text{AlexNet}_{s,\text{Corruption}}$ for each corruption type.
Gaussian Noise: 88.6\%,\quad
Shot Noise: 89.4\%,\quad
Impulse Noise: 92.3\%,\quad
Defocus Blur: 82.0\%,\quad
Glass Blur: 82.6\%,\quad
Motion Blur: 78.6\%,\quad
Zoom Blur: 79.8\%,\quad
Snow: 86.7\%,\quad
Frost: 82.7\%,\quad
Fog: 81.9\%,\quad
Brightness: 56.5\%,\quad
Contrast: 85.3\%,\quad
Elastic Transformation: 64.6\%,\quad
Pixelate: 71.8\%,\quad
JPEG: 60.7\%,\quad
Speckle Noise: 84.5\%,\quad
Gaussian Blur: 78.7\%,\quad
Spatter: 71.8\%,\quad
Saturate: 65.8\%.

\begin{table}[ht]
\footnotesize
\begin{center}
{\setlength\tabcolsep{1.1pt}%
\begin{tabular}{@{}l c ? c | c c c | c c c c | c c c  c | c c c c@{}}
\multicolumn{3}{c}{} & \multicolumn{3}{c}{Noise} & \multicolumn{4}{c}{Blur} & \multicolumn{4}{c}{Weather} & \multicolumn{4}{c}{Digital} \\
\cline{1-18}
Network & Error & \multicolumn{1}{c|}{\textbf{Rel. mCE}} & \scriptsize{Gauss.}
    & \scriptsize{Shot} & \scriptsize{Impulse} & \scriptsize{Defocus} & \scriptsize{Glass} & \scriptsize{Motion} & \scriptsize{Zoom} & \scriptsize{Snow} & \scriptsize{Frost} & \scriptsize{Fog} & \scriptsize{Bright} & \scriptsize{Contrast} & \scriptsize{Elastic} & \scriptsize{Pixel} & \scriptsize{JPEG}\\ \hline 
AlexNet    & 43.5 & 100.0 & 100 & 100 & 100 & 100 & 100 & 100 & 100 & 100 & 100 & 100 & 100 & 100 & 100 & 100 & 100 \\
SqueezeNet & 41.8 & 117.9 & 118 & 116 & 114 & 104 & 110 & 106 & 105 & 106 & 110 & 98  & 101 & 100 & 126 & 129 & 229 \\
VGG-11     & 31.0 & 123.3 & 122 & 121 & 125 & 116 & 129 & 121 & 115 & 114 & 113 & 99  & 86  & 102 & 151 & 161 & 174 \\
VGG-19     & 27.6 & 122.9 & 114 & 117 & 122 & 118 & 136 & 123 & 122 & 114 & 111 & 88  & 82  & 98  & 165 & 161 & 172 \\
VGG-19+BN  & 25.8 & 111.1 & 104 & 105 & 114 & 108 & 132 & 114 & 119 & 102 & 100 & 79  & 68  & 89  & 165 & 125 & 144 \\
ResNet-18  & 30.2 & 103.9 & 104 & 106 & 111 & 100 & 116 & 108 & 112 & 103 & 101 & 89  & 67  & 87  & 133 & 97  & 126 \\
ResNet-50  & 23.9 & 105.0 & 104 & 107 & 107 & 97  & 126 & 107 & 110 & 101 & 97  & 79  & 62  & 89  & 146 & 111 & 132 \\
\Xhline{2\arrayrulewidth}
\end{tabular}}
\caption{Relative Corruption Errors and Relative mCE values of different corruptions and architectures on \textsc{ImageNet-C}. All models are trained on clean ImageNet images, not \textsc{ImageNet-C} images. Here ``BN'' abbreviates Batch Normalization.
}
\label{tab:relarchi}
\end{center}
\end{table}

\section{10-Crop Classification Fails to Enhance Robustness}\label{appendixcrop}
\noindent Viewing an object at several different locations may give way to a more stable prediction. Having this intuition in mind, we perform 10-crop classification. 10-crop classification is executed by cropping all four corners and cropping the center of an image. These crops and their horizontal mirrors are processed through a network to produce 10 predicted class probability distributions. We average these distributions to compute the final prediction. Of course, a prediction informed by 10-crops rather than a single central crop is more accurate. Ideally, this revised prediction should be more robust too. However, the gains in mCE do not outpace the gains in accuracy on a ResNet-50. In all, 10-crop classification is a computationally expensive option which contributes to classification accuracy but not noticeably to robustness.

\section{Auxiliary Surface Variation Robustness Experiments}\label{appendixaux}
We repurpose the CIFAR-100~\citep{cifar} and ImageNet-22K datasets for a closer investigation into subtype robustness. Both datasets have hierarchical taxonomies, so they have subtypes. Then we consider an inferior but informative synthetic test for style robustness.

\noindent \textbf{CIFAR-100.}\quad The CIFAR-100 dataset has 20 broad classes, each with five subtypes. For this experiment, we hold out 1, 2, 3, or 4 subtypes from each broad class, train the classifier on the remaining subtypes while predicting the broad class, then test the classifier's broad class accuracy on the held out subtypes. The classifier is a 40-2 Wide ResNet. We find the classifier broad class prediction error rate when we hold out  1, 2, 3, or 4 subtypes, and then we average these error rates. The average broad class prediction error rate with held out subtypes as inputs is 59.7\%. In sharp contrast, the average broad class prediction error rate on subtypes known to the classifier is 13.2\%. Like before there is a dearth of subtype robustness.

\begin{wrapfigure}{r}{0.5\textwidth}
\vspace{-15pt}
  \begin{center}
      \includegraphics[width=0.48\textwidth]{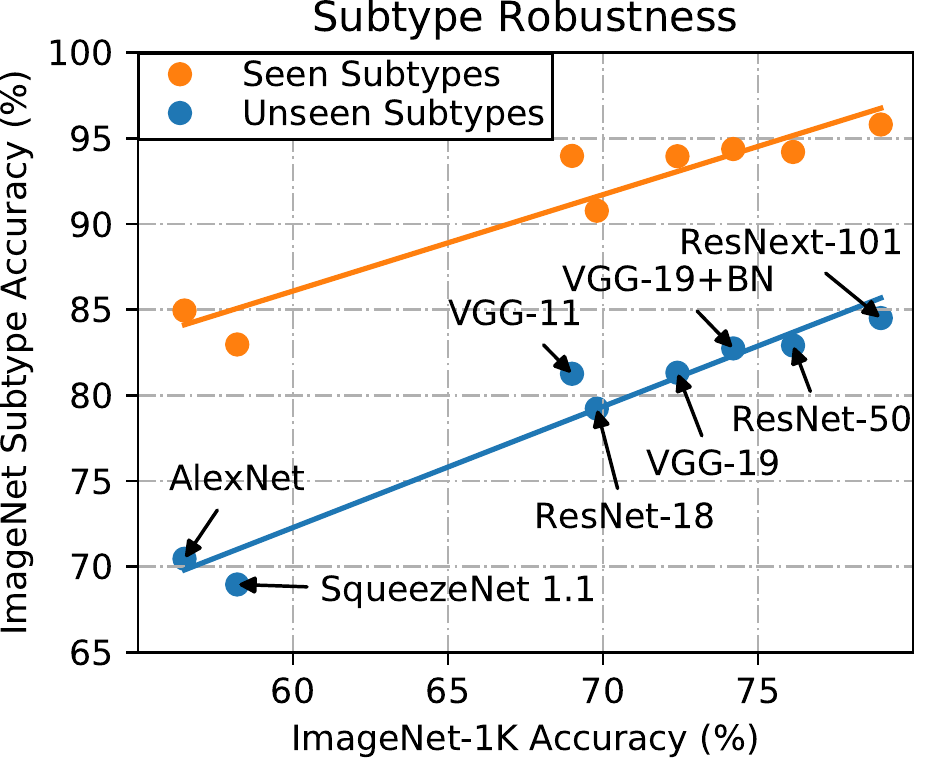}
  \end{center}
  \caption{ImageNet classifiers and their robustness to unseen subtypes. Unseen subtypes of known broad classes are noticeably harder for classifiers.}
\label{fig:superclass}
\vspace{-4pt}
\end{wrapfigure}

\noindent \textbf{ImageNet-22K.}\quad Another natural image dataset with many classes but more data is ImageNet-22K, an ImageNet-1K superset. To define this subtype robustness experiment, we manually select 25 broad classes from ImageNet-22K, listed with their WordNet IDs in the Supplementary Materials. Each broad class has many subtypes. We call a subtype ``seen'' if and only if it is in ImageNet-1K and a subtype of one of the 25 broad classes. The subtype is ``unseen'' if and only if it is a subtype of the 25 broad classes and is from ImageNet-22K but not ImageNet-1K. Fortunately, pre-trained ImageNet-1K classifiers are readily available and have not trained on subtypes which should remain unseen. Therefore, we test subtype robustness by fine-tuning several pre-trained ImageNet-1K classifiers on seen subtypes so that they predict one of 25 broad classes. Their ``seen'' and ``unseen'' accuracies are shown in Figure~\ref{fig:superclass}, while the ImageNet-1K classification accuracy before fine-tuning is on the horizontal axis. Despite only having 25 classes and having trained on millions of images, these classifiers demonstrate a subtype robustness performance gap that should be far less pronounced.

For completeness, we list the 25 broad classes which we selected from ImageNet. Amphibian (n01627424),\;
Appliance (n02729837),\;
Aquatic Mammal (n02062017),\;
Bird (n01503061),\;
Bear (n02131653),\;
Beverage (n07881800),\;
Big cat (n02127808),\;
Building (n02913152),\;
Cat (n02121620),\;
Clothing (n03051540),\;
Dog (n02084071),\;
Electronic Equipment (n03278248),\;
Fish (n02512053),\;
Footwear (n03380867),\;
Fruit (n13134947),\;
Fungus (n12992868),\;
Geological Formation (n09287968),\;
Hoofed Animal (n02370806),\;
Insect (n02159955),\;
Musical Instrument (n03800933),\;
Primate (n02469914),\;
Reptile (n01661091),\;
Utensil (n04516672),\;
Vegetable (n07707451),\;
Vehicle (n04576211).

\textbf{Style Transfer.} Now we supplement our style robustness results with a lower quality, synthetic, yet informative experiment. Style transfer \cite{style} provides a synthetic way to test style robustness by algorithmically and artificially recomposing the image in the style of another image. We test robustness to style transfer in the same way we tested robustness in our \textsc{ImageNet-C} corruption benchmark. Therefore we corrupt ImageNet validation images with the style transfer method of \citet{adain}. Images are corrupted at five different severities where higher severities preserve less content and more strictly adhere to a randomly chosen target style. Improving style transfer robustness proves to be more difficult than improving \textsc{ImageNet-C} performance. From AlexNet to ResNet-50, the mCE of \textsc{ImageNet-C} distortions went from 100\% to 76.7\%, but the Corruption Error for style transfer only decreased from 100\% to 92\%. In fact, the ResNet-50 style transfer Corruption Error is higher than all \textsc{ImageNet-C} Corruption Errors, and this holds for nearly all architectures tested. This again shows that current classifiers are not yet suited to surface variations such as stylization.

\section{Classes in \textsc{Icons-50}}\label{appendixiconsclasses}
\noindent The 50 classes of \textsc{Icons-50} are as follows: Airplane, Arrow Directions, Ball, Biking, Bird, Blade,  Boat, Books, Building, Bunny Ears, Cartwheeling, Clock, Cloud, Disk, Drink, Emotion Face, Envelope, Family, Fast Train, Feline, Flag, Flower, Footwear, Golfing, Hand, Hat, Heart, Holding Hands, Japanese Ideograph, Kiss, Lock, Mailbox, Marine Animal, Medal, Money, Monkey, Moon, Mountain, Numbers, Phone, Prohibit Sign, Star, Surfing, Tree, Umbrella, Vehicle, Water Polo, Worker, Wrestling, Writing Utensil.



\end{appendices}

\end{document}